# Extraction of handwritten areas from colored image of bank checks by an hybrid method

Sofiene HABOUBI, Samia SNOUSSI MADDOURI

*Laboratory of Systems and Signal Processing (LSTS), National Engineering School of Tunis (ENIT),*
*B.P 37 Belvédère 1002, Tunisia*
Sofiene_Haboubi@yahoo.fr
Samia.Maddouri@enit.rnu.tn

*Abstract*—One of the first step in the realization of an automatic system of check recognition is the extraction of the handwritten area. We propose in this paper an hybrid method to extract these areas. This method is based on digit recognition by Fourier descriptors and different steps of colored image processing . It requires the bank recognition of its code which is located in the check marking band as well as the handwritten color recognition by the method of difference of histograms. The areas extraction is then carried out by the use of some mathematical morphology tools.

I. INTRODUCTION

The document segmentation is a necessary step in the development of an automatic recognition system. We notice, in the state of the art, that the adaptation of document recognition system to a new type of complex document is generally difficult. When handwritten script is present in the document, the segmentation operation becomes more delicate because of the variabilities of the handwriting and irregular spacing between the writing areas.

The extraction of handwritten areas from a document is one of the major problems of segmentation. There are many methods of extraction: in [5] the separation of a text from a degraded background is made by the detection of gray level of the handwritten script and supposing that the back image is bright. In [3, 7], methods of extraction from a binary image is proposed. These binary images are obtained from documents having a superior grey level of the handwritten script. The lines are then localized by a projection method. In [1], we find an original method of location of titled lines using the partial vertical projection method; this is used for the segmentation of Arabic handwritten text. Other methods use the mathematical morphology for the recognition of printed documents [2].

The automatic system of check recognition uses generally grey leveled images for the segmentation and the extraction of handwritten areas. In [6], the handwritten script extraction is made by the detection of the main lines and the classification of the related constituents by groups. The French recognition system of bank checks A2iA start by the classification of the document. When the document is knowen, it proceeds by the extraction of every zones [8]. This method is used also in [12]. In [9], the segmentation is made on binary images where the connected components detection is the main tool of handwritten areas recognition.

We propose, in this paper, a segmentation method of Tunisian checks. This method is based on : the recognition of the check type from the code of the bank located in the marking band of the check. The recognition handwritten color by the method of difference of histograms. The extraction of handwritten areas from the physical structure of the check associated to its given bank. An improvement of the extracted area, which can over flow in certain cases the predetermined area, is made by the application of some mathematical morphology tools used to find the exact limits of the related constituents. This paper is organized in the flowing way. In section 2, we present the methods of segmentation of bank checks. In section 3, we present the Tunisian check segmentation method proposed from their color image and the handwritten extraction steps. In section 4, we propose an improvement of the extraction results in the case of the over flow of the handwritten writing. In section 5, we present the obtained results on various Tunisian check banks.

II. METHOD OF BANK CHECKS SEGMENTATION

Different techniques are used in the segmentation step of bank checks. In the following sction we give four methods.

*A. Method based on detection of the base line*

This method of detection of the handwritten areas is based on the extraction of base lines of a check. It consists in discovering the connected components by classifying each of them in separate regions. Handwritten parts are then extracted with base lines by histogram method applied on regions. A stage of classification of connected components to classify each basic line is then done. Finally, each area is separated from the others according to the associated base line which is defined by its position [6].

*B. Method based on physical structure recongnition of the check*

This probabilistic method is based on the recognition of the model bank check. It consists in extracting the image parameters which allows to the classification of the check model. Each model is described by position of these areas and its image background. After, the determination of the check model, the elimination the bottom of check is done. In result, binary images of handwritten of each area are obtained [12].

*C. Method based on recognition of the writing*

This method consists in detecting connected components of the document image according to the gray levels. It proceeds by measuring certain characteristics of these components manuscripts, in order to decide if it is a handwritten or printed writing by recognition of some words [8].

*D. Method based on detection of the connected components*

This method is based on mathematical morphology transformations by the use of physical and logical structure. It builds connected components by the applicant's morphologic transformation. It makes possible to label the connected components and to build a mask of the sought area in order to extract the literal amount area from original check. This method is applied on grey level images [4].

### III. PROPOSED METHOD

The proposed method is applied on Tunisian bank check. It requires the following steps:
- Physical and logical analyze of the check structure
- Pre-processing of the check image
- Extraction of the marking band of the check
- Recognition of the the bank code by Fourier descriptors
- Extraction of handwritten zones by difference
- Localization of each area starting from its physical site
- Improvement of areas limitation

*A. Physical and logical analyze of the check structure*

The logic design of the bank check generally and the Tunisian check in particular is standardized by the central bank as it is presented in figure 1.

The check includes two parts:
- A part witch contents the textual data and called body of the check.
- A white part called band marking zone, reserved for the magnetic encoding. It is located in the bottom and over the whole length of the check. Its width is equal to 16 mm.

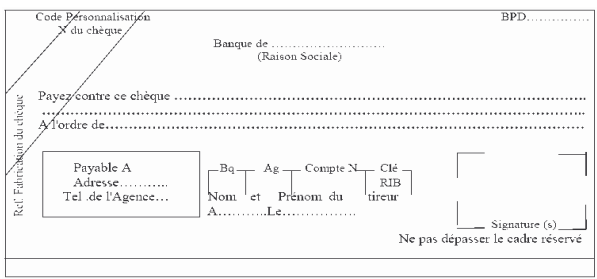

Figure 1. Model of the bank checks

*1) Body content*
The physical and logical structure is presented by the figure2.

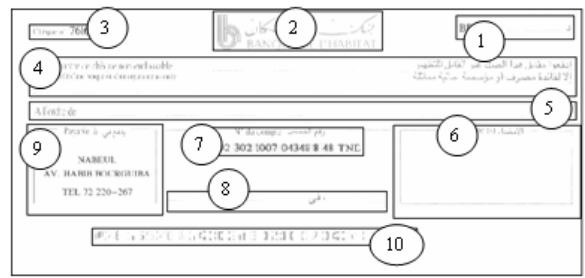

Figure 2. Physical and logical structure

- Area 1 : It contains generally printed and handwriting script. It represents the digital amount.
- Area 2 : It contains printed writing and images. It represents the name of the bank.
- Area 3 : It contains printed digits. It represents the number of the check.
- Area 4 : It contains handwriting and printed script. It represents the literal amount.
- Area 5 : It contains handwriting and printed script. It represents conductor name.
- Area 6 : It contains an image and represents the signature of the scripter.
- Area 7 : It contains printed digits. It represent the account number.
- Area 8 : It contains printed and handwriting script. It represents the date.
- Area 9 : It contains printed script and digits. It represents the address and the phone number of the bank.
- Area 10 : It contains generally digits and represents the marking band.

*2) Characteristics of the marking band*

The marking band is the second component of the check. It should contain only data codified and marked obligatorily in characters CMC7. CMC7 is a magnetic coding norm. Figure 3 presents an example of marking band of a check coming from the Tunisian Company of Banks (STB).

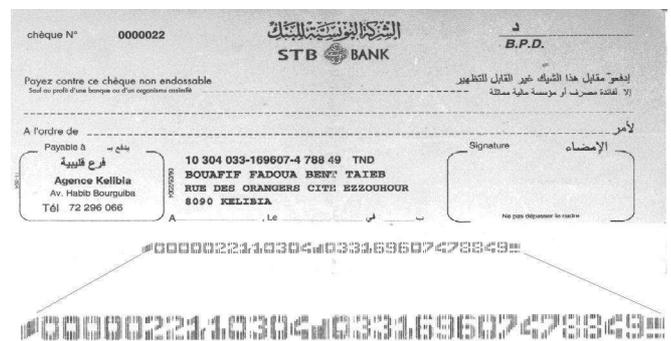

Figure 3. Bank check (STB)

Magnetic marking is comprised from the left to the right nine headings gathered into five areas.

- Area 1: this area comprises a free format of four numerical characters and the separator SI. The bank uses this area in post-printing.
- Area 2: this area comprises four headings and the separator SII, the headings are: the number of the check (seven numerical characters indicates the number of the check), nature of the account (1 numerical character), bank code (two numerical characters) and the agency codes (three numerical characters).
- Area 3: this area comprises two headings and the separator SIII. The headings are: reference of the customer account (13 numerical characters at the maximum identifying the customer account of the near the bank) and key of control of the bank account number (two numerical characters).
- Area 4: this area comprises a heading and the separator SIV. The heading is the currency code (three numerical characters).
- Area 5: this area comprises only one heading and the separator SV. The heading is check account (12 numerical characters).

The areas two and three are used obligatorily in pre-printing. The areas four and five must be post-printed. The ewistence of the first area depends on the choice of the bank.

The headings pre-printing areas constitute an identity of the banking document. There are countries which use this technique of marking with different areas and headings. From this area of marking it is interested to recognize the position and the size of the bank code in order to automatically recognize the bank and thus the check structure.

*B. Pre-processing*

This phase consists in justifying the image of a check in the global image to correct the slant and to eliminate white space on the boundary of the check. This pre-processing start by calculating the slant angular by seeking the new positions of the pixels by rotation. This step is necessary since we will extract the areas of the handwritten script by a difference between the handwritten and standard model of the bank check. The images of the two checks should have the same slant angular. Figure 4, presents an example of bank check before and after pre-processing.

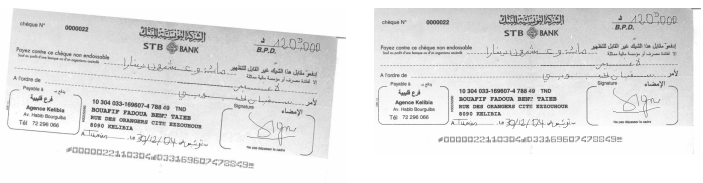

Figure 4.  Check before and after pre-processing

*C. Recognition of the code of the bank by Fourier Descriptors*

The bank recognition is done from its code on the marking band. The band of marking is composed by significant codes witch are used for banking identity. The automatic recognition system must know the position of bank code in the marking bank in order to extract and convert it into value, and recognize the corresponding bank. To recognize this value the system must pass by the following data processing sequence which is applied mainly on the marking band.

*Extraction of marking band:* the extraction of marking band from a check image is carried out from its horizontal projection. Since the marking band is written in a white area which is located in the lower part of the check, therefore the base of the last curve projection is the band width and its length is spread out over the length image. Figure 5 gives the horizontal projection of the check to distinguish the site marking band.

*Extraction of numerical characters of the bank code:* the numerical characters and the separators on the marking band are written under standard CMC7 norm. Each of them is represented by 7 stick separated by one or two spacing units. We can see in table 1 that at each character corresponds a single description in CMC7 norm. Two characters are separated by three spacing units. It is possible to recognize automatically these characters while being based on the positions of the sticks, but the noise carried out by the scanner and the small sizes of the characters as presented in the figure 6-a makes this alternative very difficult. That is why other recognition method should be used. Indeed, the flat projection of the marking band in the figure 6-b, shows that each character is represented by several connected components. Thus, we apply an ultimate dilatation until each numerical character will be constituted by only one connecteded component. The result is illustrated in the figure 6-c and in table 1.

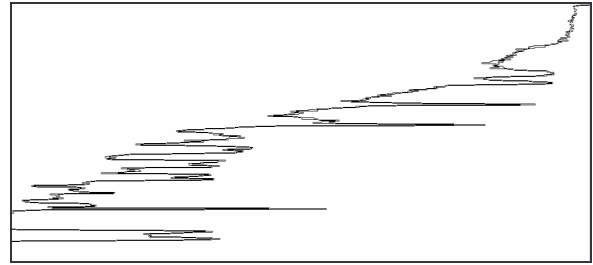

Figure 5.  Histogram of vertical projection a bank check

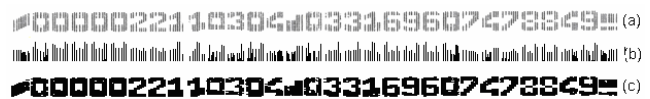

Figure 6.  (a) band marking, (b) the vertcal projection of the marking band before dilation, (c) the marking band after dilation

According to the position number 21 and 22, after a vertical projection step, of banking code. We extract the numerical characters to recugn.

TABLE I. THE CMC7 REFERENCES NUMBERS BEFORE (BD) AND AFTER DILATATION (AD)

| | 0 | 1 | 2 | 3 | 4 | 5 | 6 | 7 | 8 | 9 |
|---|---|---|---|---|---|---|---|---|---|---|
| BD | 0 | 1 | 2 | 3 | 4 | 5 | 6 | 7 | 8 | 9 |
| AD | 0 | 1 | 2 | 3 | 4 | 5 | 6 | 7 | 8 | 9 |

*Recognition of dilated numerical characters by Fourier Descriptors:* The choice of the Fourier descriptors is due to their properties of invariability, stability and completed following a simple stage of normalization. In fact, starting from the contour function of the image, we can apply Fast Fourir Transform (FFT) algorithme in order to generate set of numbers known as Fourier Descriptors or harmonics. These descriptors represent the form in frequencies field. The low frequency descriptors describe general form. The high frequency descriptors describe details. As it is shown in table 2 superposition of the N harmonics regenerate the original boundary of the image. N is the number

For classification, a subset of Fourier descriptors is often sufficient to discriminate the various forms. In table 2, from 30 harmonics we can obtain a boundary image near the original one. Since the practical applications use the discrete data, it is reasonable to employ the discrete Fourier Transformation. The equation (1) describe the Fourier transform contour function with length N, witch $A_0$ and $C_0$ are the first Fourier coefficients with order 0 and $a_n$, $b_n$, $c_n$, $d_n$ are Nth Fourier descriptors.

The approximation of the contour by Fourier coefficients can be seen like a superposition of elliptic forms with different sizes, positions and orientations. A step of normalisation of start point of the contour, size, position and orientation is applied.

$$X_N(k) = A_0 + \sum_{n=1}^{N} a_n \cos \frac{2n\pi k}{N} + b_n \sin \frac{2n\pi k}{N}$$

$$Y_N(k) = C_0 + \sum_{n=1}^{N} c_n \cos \frac{2n\pi k}{N} + d_n \sin \frac{2n\pi k}{N}$$ (1)

Following the standardization of the Fourier Descriptors (FD) of either the dilated numerical characters or reference numbers, a calculation of distance is carried out between the FD of numerical characters to recognize and each of the reference numbers. This distance is defined by the equation (2) where "cd", the dilated numerical characters and "rf", the reference numbers [10].

$$D(cd, rf) = \frac{\sum_{i=1}^{N}(a_{cd}^2 - a_{rf}^2) + (b_{cd}^2 - b_{rf}^2) + (c_{cd}^2 - c_{rf}^2) + (d_{cd}^2 - d_{rf}^2)}{\sum_{i=1}^{N}(a_{cd}^2 + a_{rf}^2 + b_{cd}^2 + b_{rf}^2 + c_{cd}^2 + c_{rf}^2 + d_{cd}^2 + d_{rf}^2)}$$ (2)

TABLE II. FOURIER APPROXIMATION, OF THE BOUNDARY OF THE CMC7 NUMBER 3 AFTER DILATATION BY ADDITION OF 30 HARMONICS

| 1 | 2 | 3 | 4 |
|---|---|---|---|
| 0 | ⊃ | 3 | 3 |
| 6 | 8 | 10 | 12 |
| 3 | 3 | 3 | 3 |
| 15 | 20 | 25 | 30 |
| 3 | 3 | 3 | 3 |

D. *Extraction of handwritten areas*

This step initially consists in separating the handwritten from the background of the image and to extract each area independently from the others. Thus the system must pass by the two following steps:

- Step 1: we make the difference between the histogram of the check filled of figure 4 and the histogram of the check not filled. The later is already known from the bank code of its. The histogram used is the number of pixels for each colour of a pallet RGB. The maximum peak of the difference represents the colour of handwriting. We apply then a passes band filters to obtain only handwriting. The result is illustrated by the figure 7-a.

- Step 2: each bank has a model of check carrying a well defined structure. This structure describes the position of each area. These areas are then extracted in separate images since the model is recognized by its banking code. .In table 3 we find exemples of positions of some area for three tunisien bank ftoken from our database. The figure 7-b presents the limit of each handwritten area.

TABLE III.  EXEMPLES OF STANDARD POSITION OF SOME AREA OF TWO TUNISIEN BANCK (BFT AND STB)

|     |   | MN  | ML   | Ben. | Date | Sign. |
|-----|---|-----|------|------|------|-------|
| BFT | X | 866 | 0    | 0    | 266  | 723   |
|     | Y | 42  | 110  | 193  | 344  | 244   |
|     | L | 175 | 1043 | 1043 | 438  | 326   |
|     | H | 56  | 83   | 40   | 53   | 132   |
| STB | X | 758 | 0    | 0    | 240  | 675   |
|     | Y | 22  | 116  | 198  | 347  | 249   |
|     | L | 226 | 974  | 974  | 412  | 288   |
|     | H | 60  | 89   | 44   | 38   | 128   |

where:
(X,Y) : position of the first point in the top and in the left.
MN: Numerical amount zone
ML: Literal amount zone
Ben: Conductor name zone
Date: Date zone
Sign: Signature zone

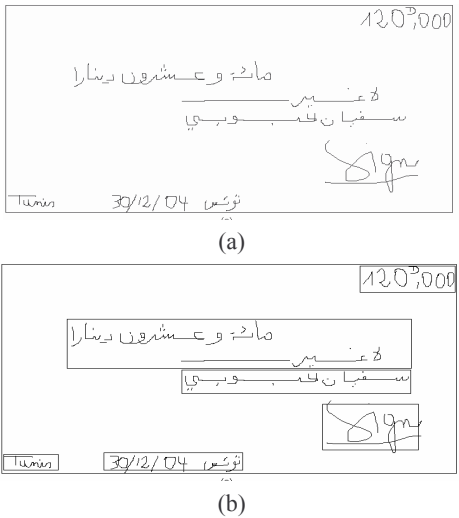

(a)

(b)

Figure 7. (a) handwritten areas extracted after method of difference, (b) handwritten delimited areas after recognition code bank

Although handwritten areas are known and bounded the user can exceed the area which is reserved for him, such as the case of the signature in the figure 7-b. Also, extracted areas are composed by more then one connected components. An improvement step is given in the flowing section to limitation area.

IV. IMPROVEMENT OF HANDWRITTEN AREA LIMITATION

In some cases, the handwritten can overflow from its reserved area. We chose to change the limits of area presenting intersection with the handwriting boundary. The enlarging consists in displacing the superior, lower, left and right limits of area so much that they do not cut the handwritten script. The result is illustrated in figure 8.

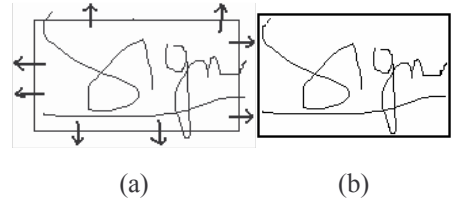

(a)           (b)

Figure 8. (a) Before improvement, (b) After improvement

In other case, we found in extracted areas more the one connected components. Second proposed improvement is the extraction of all connected components associated to each area.

It is easy to note that the Arab handwriting is coposed by connected components corresponding to pieces of words known as PAW: "Pieces of Arabic Word". To connect them it is not enough to apply dilation by an horizontal segment as structuring element. Dilation by a vertical segment could connect the diacritic points and the adjacent lines. The operation of dilation is applied on image of the figure 9-a. the result is illustrated by the figure 9-b.

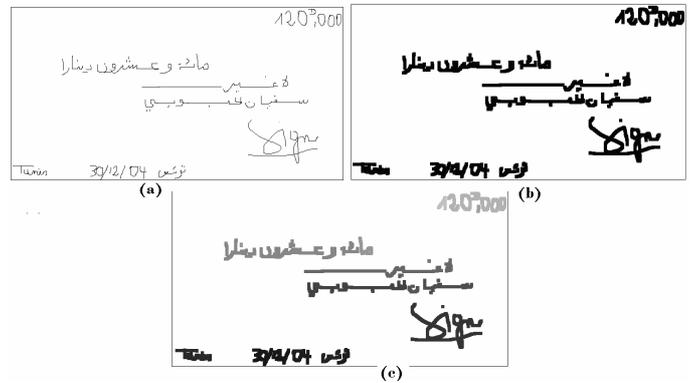

Figure 9. (a) binary image of manuscript, (b) vertical horizontal dilation on level 3, (c) labelling of the related components

Once the related component is delimited, a step of labelling is necessary in this phase for the distinction between each component. The principle of this stage is to assign to each area a label corresponding to a colour. This treatment is applied to the image of figure 9-b. the result is shown by figure 9-c.

The extraction of connected components of each area is applied on the image of figure 9-c after labelling. We associate each connected components to its associated area by a variation of that all pixels of the later belong to the set pixels inside the former surface. Figure 10 presents the result of this step. We can see that digital amount area is composed by one connected component, however literal amount area contains 9 connected components. These connected presented in figure 11 would be presented to the recognition system of handwritten Arabic PAWs.

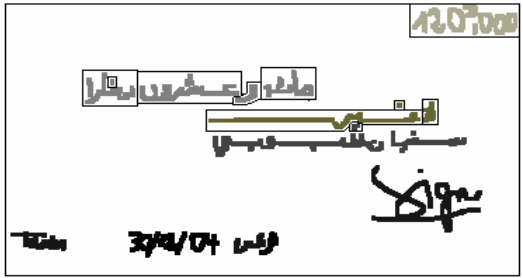

Figure 10. Extraction of connected components

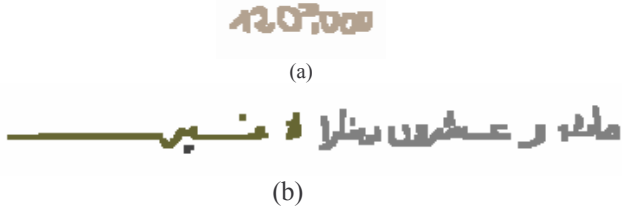

(a)

(b)

Figure 11. Result of the extraction of numerical (a)and literal (b) amount

## V. EXPERIMENTAL RESULTS

To evaluate our method, we have used six different models of Tunisians bank checks. 20 checks are filled for every model. The basis contains 120 checks. This reduced number is due to the fact that real database are not available because of the confidentiality. We find in the table 1, the results of assessment of the step of separation of the numbers. 7 banking codes didn't separate correctly, because of the small spaces that separate the numbers, this increase the risk to have complete numbers.

TABLE IV. RATE OF SEPARATION

| Total number of checks | NCCSC | NCISC | Rate separation | Rate error |
|---|---|---|---|---|
| 120 | 113 | 7 | 94% | 6% |

NCCSC: Number of Checks having Correctly Segmented Codes
NCISC: Number of Checks having Incorrectly Segmented Codes

The formula (2) of distance proved its efficiency as well for

In the table 4, we give the results of distance between the FD references numbers and numbers ones extracted from marking band after a step of dilation. We can see that the diagonal is always in the minimum.

In the table 5, we present the recognition of the codes of Tunisian banks. We notice that the minimal distance (in grey) is always associated to the adequate number of the code of the bank. It gives us a rate of recognition of 100% for the 6 band checks

TABLE V. CALCULATION OF THE DISTANCES BETWEEN THE FD REFERENCES NUMBERS AND THE FD NUMBERS EXTRACTED FROM MARKING BAND OF SOME BANK CHEKS

| References extracted | 0 | 1 | 2 | 3 | 4 | 5 | 6 | 7 | 8 | 9 |
|---|---|---|---|---|---|---|---|---|---|---|
| 0 | 0.0093 | 0.0302 | 0.0528 | 0.0143 | 0.0305 | 0.0794 | 0.0138 | 0.0154 | 0.1924 | 0.0445 |
| 1 | 0.0148 | 0.0039 | 0.0041 | 0.0242 | 0.0112 | 0.0135 | 0.0306 | 0.0288 | 0.3717 | 0.0102 |
| 2 | 0.0229 | 0.0150 | 0.0021 | 0.0280 | 0.0154 | 0.0078 | 0.0372 | 0.0361 | 0.3897 | 0.0104 |
| 3 | 0.0097 | 0.0050 | 0.0113 | 0.0022 | 0.0076 | 0.0282 | 0.0096 | 0.0103 | 0.2992 | 0.0136 |
| 4 | 0.0083 | 0.0093 | 0.0146 | 0.0135 | 0.0021 | 0.0302 | 0.0172 | 0.0107 | 0.3222 | 0.0182 |
| 5 | 0.0354 | 0.0262 | 0.0048 | 0.0353 | 0.0210 | 0.0007 | 0.0434 | 0.0440 | 0.4001 | 0.0086 |
| 6 | 0.0111 | 0.0079 | 0.0228 | 0.0070 | 0.0085 | 0.0420 | 0.0008 | 0.0022 | 0.2493 | 0.1665 |
| 7 | 0.0380 | 0.0089 | 0.0169 | 0.0097 | 0.0057 | 0.0308 | 0.0037 | 0.0034 | 0.2644 | 0.0110 |
| 8 | 0.1218 | 0.1117 | 0.1605 | 0.0854 | 0.1148 | 0.7985 | 0.0726 | 0.0775 | 0.1056 | 0.1435 |
| 9 | 0.0186 | 0.0128 | 0.0031 | 0.0171 | 0.0112 | 0.0083 | 0.0189 | 0.0218 | 0.3237 | 0.0004 |

TABLE VI. RECOGNITION OF SOME SEPARATE CODES

|   | BFT | | BNA | | BS | | STB | | UIB | | BH | |
|---|---|---|---|---|---|---|---|---|---|---|---|---|
|   | *0* | *2* | *0* | *3* | *0* | *4* | *1* | *0* | *1* | *2* | *1* | *4* |
| 0 | 0.021 | 0.023 | 0.035 | 0.008 | 0.059 | 0.011 | 0.009 | 0.048 | 0.005 | 0.021 | 0.127 | 0.045 |
| 1 | 0.121 | 0.014 | 0.142 | 0.006 | 0.116 | 0.005 | 0.003 | 0.149 | 0.001 | 0.013 | 0.021 | 0.127 |
| 2 | 0.178 | 0.001 | 0.187 | 0.014 | 0.153 | 0.010 | 0.005 | 0.203 | 0.009 | 0.001 | 0.552 | 0.450 |
| 3 | 0.116 | 0.025 | 0.130 | 0.002 | 0.099 | 0.018 | 0.020 | 0.130 | 0.011 | 0.019 | 0.053 | 0.071 |
| 4 | 0.137 | 0.015 | 0.154 | 0.004 | 0.125 | 0.002 | 0.007 | 0.161 | 0.007 | 0.015 | 0.109 | 0.014 |
| 5 | 0.211 | 0.006 | 0.222 | 0.031 | 0.186 | 0.021 | 0.015 | 0.245 | 0.027 | 0.006 | 0.211 | 0.076 |
| 6 | 0.097 | 0.034 | 0.109 | 0.005 | 0.083 | 0.021 | 0.023 | 0.141 | 0.014 | 0.027 | 0.154 | 0.427 |
| 7 | 0.106 | 0.034 | 0.124 | 0.004 | 0.097 | 0.015 | 0.021 | 0.124 | 0.014 | 0.306 | 0.124 | 0.176 |
| 8 | 0.057 | 0.381 | 0.053 | 0.285 | 0.111 | 0.339 | 0.346 | 0.048 | 0.318 | 0.364 | 0.088 | 0.448 |
| 9 | 0.151 | 0.008 | 0.163 | 0.013 | 0.132 | 0.014 | 0.008 | 0.179 | 0.013 | 0.004 | 0.361 | 0.772 |

In the table 6, we present the results of the extraction of the handwritten areas. We give for every kind of area the total number of areas in the test basis and the number of areas correctly extracted. The total extraction rate is estimated to 95%.The total rate of mistakes is estimated to 5%. It is due to the fact that in some areas the writing overflows. We notice that this overflow is a lot more frequent in the signature area, or in the literal amount areas and the conductor's name witch generate in most cases an overlapping between these two areas. In the figure 12, we give an example of check containing these two types of overlapping.

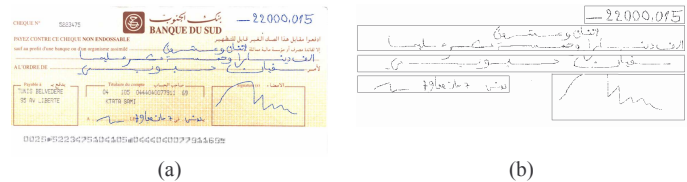

(a)          (b)

Figure 12. example of overflow

TABLE VII. RATE OF EXTRACTION OF THE HANDWEITTEN AREA

|   | SDA[1] | SLA[2] | SC[3] | SD[4] | SS[5] | TSA[6] |
|---|---|---|---|---|---|---|
| CA[7] | 120 | 114 | 114 | 120 | 107 | 575 |
| Rate extraction | 100% | 94% | 94% | 100% | 89% | 95% |
| Rate error | 0% | 6% | 6% | 0% | 11% | 5% |

[1] Segmented Digital Amount
[2] Segmented Literal Amount
[3] Segmented Conductor
[4] Segmented Date
[5] Segmented Signature
[6] Total of Segmented Areas
[7] Counts of Areas

In the table 7, we estimate the improvement rate of the extraction module. This rate varies between 3% and 11%. The error rate of 3%, obtained after the improvement, is associated to the two zones of the literal amount and the conductor's name. The enlarging of the areas doesn't manage to separate all cases. Some cases of overlap require more stages of characters recognition and segmentation.

TABLE VIII. RATE OF EXTRACTION OF THE HANDWRITTEN AREAS ATFTER IMPROVEMENT

|   | SDA | SLA | SC | SD | SS | TSA |
|---|---|---|---|---|---|---|
| Extraction rate before Improvement. | 100% | 94% | 94% | 100% | 89% | 95% |
| Extraction rate after Improvement. | 100% | 97% | 97% | 100% | 100% | 99% |
| Improvement Rate | 0% | 3% | 3% | 0% | 11% | 4% |

VI. CONCLUSION

We present in this paper an hybrid method of extraction of handwritten areas of Tunisien bank check based on image processing and numerical pattern recognition. After the recognition of its model by the use of bank code in the marking band of, the colour of the handwritten is detected by difference between the histograms colours of the model check and the filled check. The research of connected components checks limits of areas chosen according to the model and exchanges an overflow case. Lastly, the separation of each area is independent from others.

The rates of extraction are varied from a module to other. The evaluation of the recognition module of the banking code gives a rate to 94%, 6% of errors are essentially due to the existence of other writings that figures on the marking band or deformation in the banking document. The module of handwritten areas extraction gives a rate to 95% before the improvement of areas limits, and 99% after improvement with a variation between 3% and 11%, the main cause of error is overlapping between the name of the conductor and the literal amount.

In prospect we propose to introduce modules of recognition to be able to distinguish between the various areas in overlapping. The second prospect under development is the classification of the zones automatically extracted in a data base for later uses recognition of literal, rising amounts numerical or signatures.

REFERENCES

[1] A. Bennasri, A. Zahour and B. Taconet. extraction des lignes d'un texte manuscrit arabe. *Vision interface, pp 41-47,1999.*

[2] B. Al-Badr and R. Haralick. Recognition without segmentation: using mathematical morphology to recognise printed Arabic. *13th National Computer Conference, Riyadh, Saudi Arabia, pp. 813--829, November.*

[3] B. Waked, S. Bergler, C. Suen and S. Khoury. Skew detection, pages segmentation, and script classification of printed document images. *IEEE International Conference on Systems, Man, and Cybernetics (SMC'98), pp. 4470-4475,1998.*

[4] F. Bouafif Samoud, S. Snoussi Maddouri and N. Ellouze. Automate extraction and segmentation of literal and digit amount from Arabic checks. *Third International Conference on Signals, Systems, Devices.* pp.154. 2005.

[5] G. Leedham, S. Varma, A. Patankar and V. Govindaraju. Separating text and background in degraded document images - A comparaison of global thresholding techniques for multi-stage thresholding. *Proceedings Eighth International Workshop on Frontiers of Handwriting Recogni-tion, pp. 244-249, , 2002.*

[6] K. Liu and C. Suen and C. Nadal. Automatic extraction of items from cheque images for payment recognition. *In Proceedings of the International Conference on Pattern Recognition, pp 798-802, 1996.*

[7] M. Jung and Y. Shin and S. Srihari. Machine printed character segmentation method using side profiles. *Proceedings IEEE International Conference on Systems, Man and Cybernetics, 1999.*

[8] N. Gorski , V. Anisimov, E. Augustin, O. Baret and S. Maximov. Industrial bank check processing: the A2iA CheckReader. *pp 196-206, 2001.*

[9] R. Palacios and A. Gupta. A system for processing handwritten bank checks automatically. *Image andVision Computing, Feb 2002.*

[10] S. Snoussi.Maddouri, H. Amiri, A. Belaid and C. Choisy, Combination of Local and Global Vision Modeling for Arabic Handwritten Word Recognition, *International Workshop Frontier in HandWriting IWFHR'02, Canada, 2002.*

[11] S. Snoussi Maddouri, A. Belaid, Ch. Choisy and H. Amiri. Modèle perceptif neuronal à vision globale-locale pour la reconnaissance de mots manuscrits arabes. *Conférence Internationale Francophone sur l'Ecrit et le Document - CIFED 2002, Hammamet, Tunisie. 2002. 15.*

[12] Y. Al-Ohali, M. Cheriet and C. Suen. Databases for handwritten arabic checks. *Pattern Recognition 36 pp 111 – 121,2003.*